\documentclass{article}

\PassOptionsToPackage{square,numbers}{natbib}
\usepackage[final]{neurips_2023}

\usepackage[utf8]{inputenc} % allow utf-8 input
\usepackage[T1]{fontenc}    % use 8-bit T1 fonts
\usepackage{hyperref}       % hyperlinks
\usepackage{url}            % simple URL typesetting
\usepackage{booktabs}       % professional-quality tables
\usepackage{amsfonts}       % blackboard math symbols
\usepackage{nicefrac}       % compact symbols for 1/2, etc.
\usepackage{microtype}      % microtypography
\usepackage{xcolor}         % colors
\usepackage{graphicx}

\title{Fewshot learning on global multimodal embeddings for earth observation tasks}

\author{%
  Matt Allen \\
  University of Cambridge, UK\\
  \texttt{mja78@cam.ac.uk} 
  \And
  Francisco Dorr \\
  Independent, Argentina\\
  \texttt{fran.dorr@gmail.com} 
  \And
  Joseph A. Gallego-Mejia \\
  Universidad Nacional de Colombia, Colombia\\
  \texttt{jagallegom@unal.edu.co} 
  \And
  Laura Martínez-Ferrer \\
  Universitat de Val\`encia, Spain\\
  \texttt{laura.martinez-ferrer@uv.es} 
  \And
  Anna Jungbluth \\
  European Space Agency, Climate Office, UK\\
  \texttt{anna.jungbluth@esa.int} 
  \And
  Freddie Kalaitzis \\
  University of Oxford, UK\\
  \texttt{freddie.kalaitzis@cs.ox.ac.uk} 
  \And
  Raúl Ramos-Pollán \\
  Universidad de Antioquia, Colombia\\
  \texttt{raul.ramos@udea.edu.co}
}

\begin{document}

\maketitle

\begin{abstract}
In this work we pretrain a CLIP/ViT based model using three different modalities of satellite imagery across five AOIs covering over ~10\% of Earth's total landmass, namely Sentinel 2 RGB optical imagery, Sentinel 1 SAR radar amplitude and interferometric coherence. This model uses $\sim 250$ M parameters. Then, we use the embeddings produced for each modality with a classical machine learning method to attempt different downstream tasks for earth observation related to vegetation, built up surface, croplands and permanent water. We consistently show how we reduce the need for labeled data by 99\%, so that with ~200-500 randomly selected labeled examples (around 4K-10K km$^2$) we reach performance levels analogous to those achieved with the full labeled datasets (about 150K image chips or 3M km$^2$ in each area of interest - AOI) on all modalities, AOIs and downstream tasks. This leads us to think that the model has captured significant earth features useful in a wide variety of scenarios. To enhance our model's usability in practice, its architecture allows inference in contexts with missing modalities and even missing channels within each modality. Additionally, we visually show that this embedding space, obtained with no labels, is sensible to the different earth features represented by the labelled datasets we selected.
\end{abstract}

\section{Introduction}
Earth Observation (EO) has made remarkable progress with the rise of deep learning (DL) methods in recent years, and this potential is fueled by the ever increasing  availability of satellite imagery \cite{zhu2017deep}. According to the UCS Satellite Database\footnote{https://www.ucsusa.org/resources/satellite-database}, as of May 2022 there were 470 optical satellites in orbit, 102 radar satellites and 62 tagged as producing some form of imaging (hyperspectral, multispectral), among others. Within ESA's Sentinel missions alone, 80 PB of user-level data were downloaded during 2021 \cite{copernicus2021}.

However, it is well known that DL methods are overwhelmingly hungry for labeled data, and one of the main hurdles to effectively exploit DL methods in EO is its endemic scarcity \cite{wang2022self}. Even if new EO annotated datasets are published regularly, the time and effort involved cannot keep up with the amount of new data produced by present and future orbiting missions. It is in this context that methods that can significantly reduce the need for labeled data become valuable. See Section \ref{sec:prevworks}.

CLIP \cite{radford2021learning} is an SSL method that constrastively learns how to align the representations of image/text pairs collected from the Internet, allowing it to deal with different modalities of the same reality within the same embedding space for a variety multi-modal applications including detection, captioning, VQA and conditional image generation among others. CLIP's architecture, heavily based on Vision Transformers (ViT) \cite{dosovitskiy2021image}, has been applied to merge multimodal representations beyond text and in different domains \cite{xue2022clip}, \cite{wang2022medclip}, \cite{li2021align}. CLIP like models are also starting to appear in satellite imagery under different scenarios for temporal and multi-spectral imagery \cite{cong2022satmae},  temporal multimodal pixel wise \cite{tseng2023lightweight} or a multispectral contrastive model for Landsat imagery \cite{stewart2023ssl4eo}.

In this work we built a three tower CLIP architecture, feed them with Sentinel 2 RGB optical imagery, Sentinel 1 amplitude and Sentinel 1 inferometric coherence, and use the produced embeddings in several downstream classification tasks, representing different earth features. We show how, in this representation space, a small fraction of data is enough to obtain full-dataset level performance, reducing by two orders of magnitude the need for labeled data.

This paper is structured as follows. Section \ref{sec:prevworks} discusses related previous works. Section \ref{sec:data} describes the Areas of Interest (AOI) used, the input imagery and downstream labels. Section \ref{sec:method} describes the SSL architecture and training procedure, together with the downstream task. Section \ref{sec:results} shows results and visualization and in Section \ref{sec:conclusion} we draw some conclusions.

\section{Previous works}
\label{sec:prevworks}
A number of approaches have been developed to address the labeled data scarcity challenge, including a variety of methods under self supervised learning (SSL) \cite{wang2022self} and weakly supervised methods \cite{fasana2022weakly}, among others, more often that not blended into foundation models \cite{lacoste2021toward}.  Weakly supervised methods consider a range of scenarios where labels are noisy, incomplete, inexact or inaccurate and have also been applied in Earth Observation \cite{yue2022optical}. For instance, the teams on the data fusion contest \cite{robinson2021global} attempt to produce fine grained semantic segmentation maps for land cover when only low resolution reference data is available. Or also, in \cite{wang2022unlocking} a transfer learning method is used to pre-train a model over a region with large label density, to finetune it somewhere else with very few labels. 

The success of foundation models in language tasks is still hard to translate to earth observation scenarios \cite{mai2023opportunities}, but there are convincing works pre-training models with large amounts of geospatial data with the expectation to be useful in a wide variety of downstream tasks, see \cite{jakubik2023foundation}, \cite{sun2022ringmo} or \cite{wang2022advancing}. Our work contributes in such direction by explicitly considering multimodality in the pretraining step while allowing downstream applications to use function even if not all modalities or channels are available. Given the variability of EO data across time and geographical locations, we believe this is a key step to enhance the practical applicability of general pretrained models in EO.

\section{Data and downstream tasks}
\label{sec:data}

\paragraph{Input modalities}

We use three modalities for input data, taking observations during the first three months of 2020, obtained from the Sentinel-1 and Sentinel-2 ESA missions. Sentinel-1 is a Synthetic Aperture Radar (SAR) sensor, for which we use amplitude and coherence, whereas Sentinel-2 is an optical satellite. The intuition here is that both missions complement each other, offering different perspectives on the same earth features which are necessarily correlated. For each AOI (see below) we built a grid containing tiles (image chips) of size 448m $\times$ 448m. For Sentinel-2 optical data (\texttt{s2rgbm}) we use only the three RGB channels which we average per month (thus, 9 channels). For Sentinel-1 SAR amplitude (\texttt{s1grdm}) we use the vv and vh polarizations, plus their logarithmic difference, taking also their average per month (thus, 9 channels). Both Sentinel-2 and Sentinel-1 amplitude were tiled from Google Earth Engine using \texttt{geetiles}\footnote{\url{https://github.com/rramosp/geetiles}} which provides a 10m resolution and thus each chip is 448x448 pixels. We obtained Sentinel-1 interferometric coherence (\texttt{gunw}) from ARIA Sentinel-1 Geocoded Unwrapped Interferograms database \cite{buzzanga2020toward} as available through Alaska's Satellite Facility\footnote{\url{https://asf.alaska.edu/data-sets/derived-data-sets/sentinel-1-interferograms/}}, using \texttt{sartiles}\footnote{\url{https://github.com/rramosp/sartiles}}. Since interferometric coherence is built using pairs of Sentinel-1 observations we selected the pairs available whose second observation was within 2020Q1 and the first one at most 48 days before. Thus, we have a potentially variable number of channels in each tile, depending on the number of interferometric pairs which could be formed. Its resolution is around 90m per pixel, and we upsample the image chips to match the 448x448 pixel size of \texttt{s1grdm} and \texttt{s2rgbm}.

\begin{figure}[h]
\centering
\caption{Areas of Interest (AOIs) used in this study. Bands indicate the splits for train (yellow), validation (blue) and test (pink). In total there are 167K image chips for CONUS, 163K chips for Middle East, 147K chips for Pakistan-India, 285K chips for China and 83K chips for South America, which aggregates to 845K chips covering a surface of 16.9M km2.
\label{fig:aois}}
\centering
\includegraphics[width=1.0\textwidth]{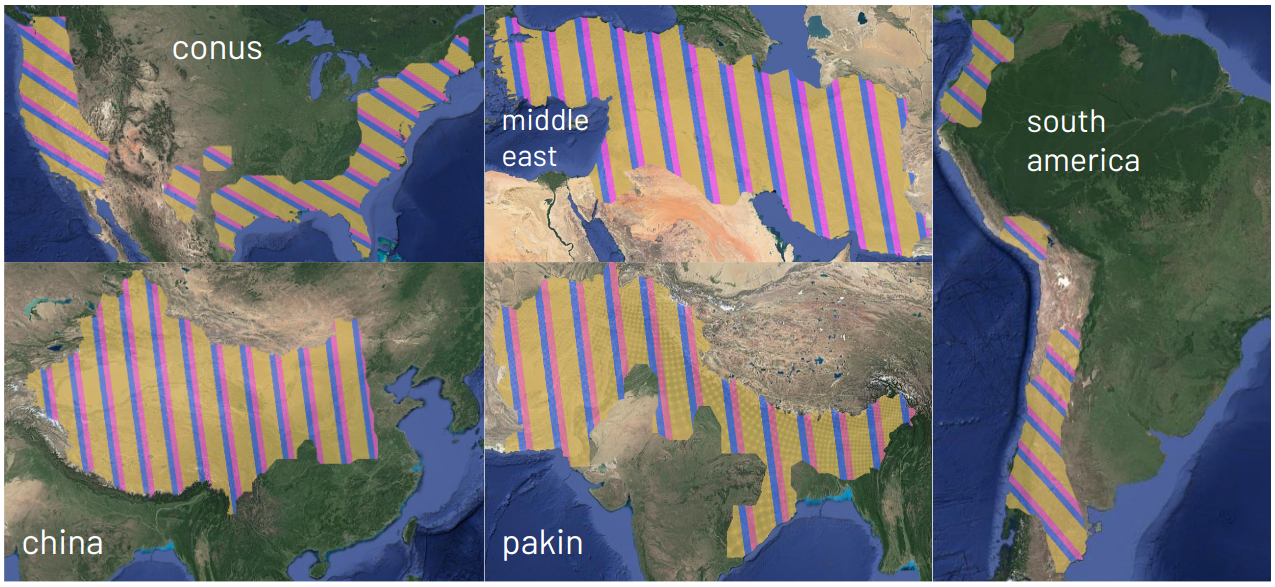}
\end{figure}

\paragraph{Areas of Interest}

We defined five AOIs covering regions in Conterminous United States (CONUS, 167K image chips), South America (83K chips), Pakistan and India (Pakin, 147K chips), China (285K chips) and the Middle East (163K chips), as illustrated in Fig. \ref{fig:aois}. The AOIs extent is determined by the 2020 coverage of the \texttt{gunw} dataset. Observe that we do geographical aware splits into train (60\%), validation (20\%) and test (20\%) to avoid as much as possible data leakage from contiguous image chips being in different splits.

\paragraph{Downstream tasks}
We selected four use cases with global label coverage so that we could experiment on ablations with an increasing number of available labels. \textbf{Vegetation estimation}: we used the MOD44B.006 Terra vegetation continuous fields yearly dataset for 2020, focusing on the tree percentage estimation at 250m per pixel\footnote{\url{https://developers.google.com/earth-engine/datasets/catalog/MODIS_006_MOD44B}}.  \textbf{Built Up Surface}, the Global Human Settlement Layer Built-Up surface dataset from the Joint Research Center of the European Commission\footnote{\url{https://ghsl.jrc.ec.europa.eu/download.php?ds=bu}} for 2020 at 100m per pixel. \textbf{Croplands}, the ESA World Cover 2020 class representing croplands \cite{zanaga_daniele_2021_5571936} at a 10m/pixel resolution. \textbf{Permanent water}, the ESA World Cover 2020 class representing permanent water bodies \cite{zanaga_daniele_2021_5571936} at a 10m/pixel resolution. The JRC dataset was downloaded from their site and tiled using \texttt{sartiles}, whereas the rest were downloaded and tiled from Google Earth Engine using \texttt{geetiles}

\begin{figure}[h]
\centering
\caption{Distribution of labels on each downstream task and AOI shown as a quantile plot. Observe that most tiles do not contain built surface or permanent waters 
\label{fig:labels-distribution}}
\centering
\includegraphics[width=1.0\textwidth]{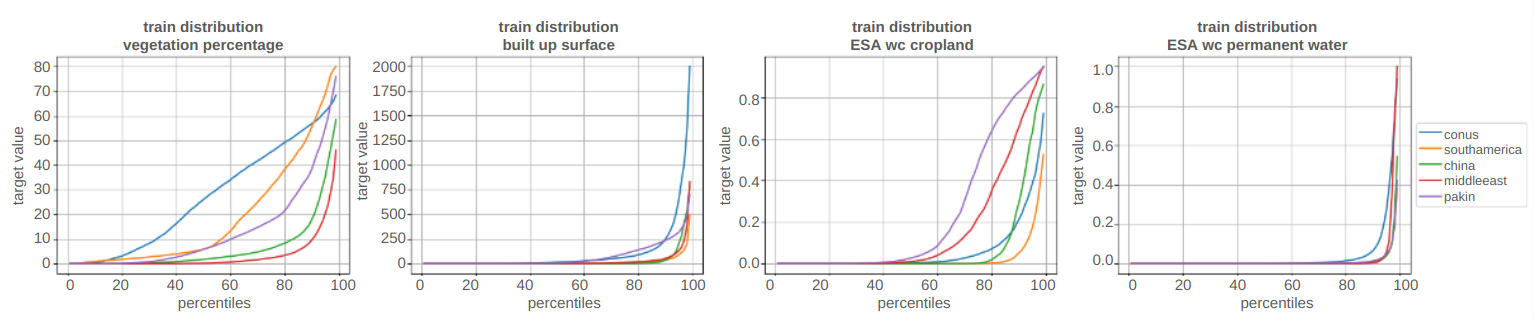}
\end{figure}

For each dataset we define a binary classification task to predict the mean value per chip, thresholded on each AOI so that we get two balanced classes. Within the same task, this threshold is usually different for each AOI as they have different distributions as shown in Fig \ref{fig:labels-distribution}. So, for instance, for \textbf{vegetation percentage} we set forth to predict whether an image chip has high or low vegetation, for \textbf{builtup} surface we predict whether a chip has high or low built surface, etc.

\section{Method}
\label{sec:method}

\subsection{Model pretraining}
\begin{figure}[h]
\centering
\caption{Architecture of our CLIP-based model with three input modalities and separate ViT encoder for each modality. Similarity is measured for each pair of modalities and then averaged. Like the original CLIP, within the same batch, our loss encourages similarity of different modalities of the same locations to have similar encodings, and from other locations to be different. Observe as well how our encoders are single channel, operating on whatever channel was randomly selected for each modality.
\label{fig:model}}
\centering
\includegraphics[width=1.0\textwidth]{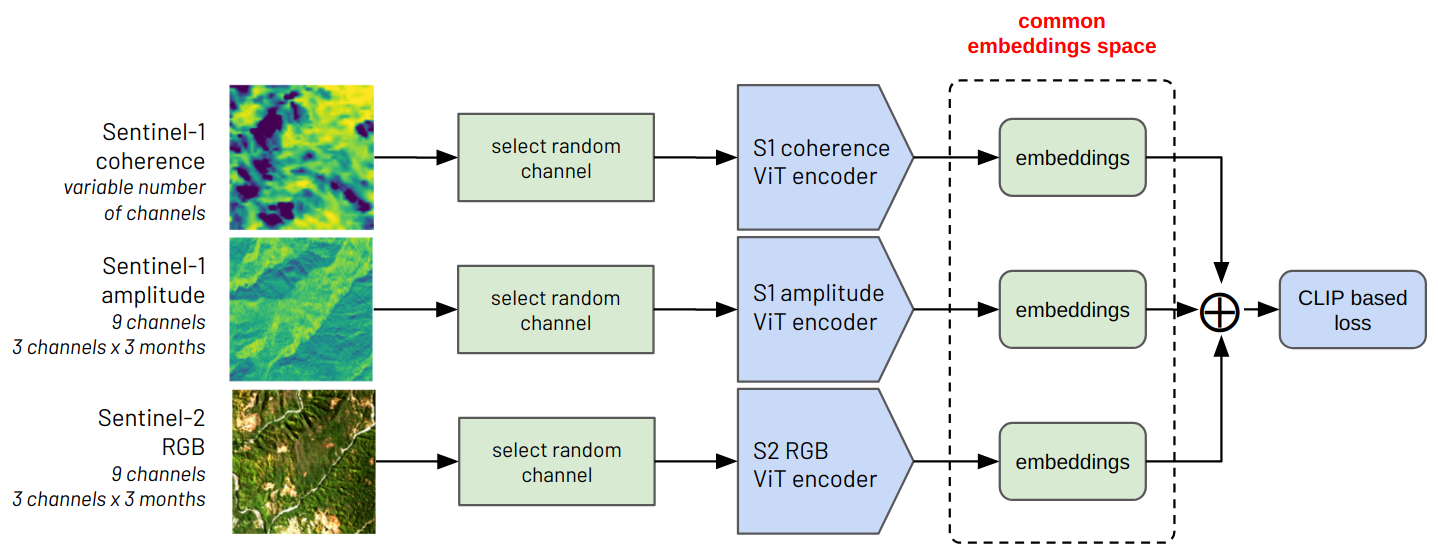}
\end{figure}

We use a Self Supervised Learning approach using Visual Transformers (ViT) and a CLIP based loss architecture, where we have one ViT tower per input modality (\verb+s1grdm+, \verb+s2rgbm+ and \verb+gunw+). This architecture produces an embedding for each input image and modality and pushes embeddings of different modalities on the same chip to be similar, and others to be different. See Figure \ref{fig:model}. In practice we are bound to occasional unavailability of particular channels: sometimes vv or vh are not available on Sentinel 1, clouds occasionally hinder Sentinel 2, and the number of interferometric pairs formed for SAR coherence is not always the same. To cope with this, each of the ViT accepts a single channel structure, and we select randomly one single channel of each input modality to feed each ViT at each inference request or training step. Besides enhancing the usability of the pretrained model for downstream tasks to noisy or missing data scenarios, we observed that this setup produces more robust outputs, probably due to the input noise induced by this procedure being compensated by correlations between modalities.

Since we tried different ViT configurations and embedding sizes, we use the train data split for training the model and the validation split to select the best self supervised model according to the CLIP loss. Test data is only using to measure the results presented here. Our final ViT configuration produces an embedding of size 768 for each input image chip in each modality, containing 259M trainable parameters. Train data amounts to about 500K image chips, equivalent to some 10M km$^2$ and taking about 100 hours of training time on an Nvidia DGX architecture with 8 A100 GPUs, 250 CPU cores and 2TB RAM.

\subsection{Downstream tasks}
For each of the five AOIs and each modality we train a Random Forest to binary classify whether the mean value of each measure (vegetation, built surface, croplands and permanent water) is below or above the median. We use the embeddings representation for each modality as produced by the pretrained model which have a size of 768 components. We create an additional modality by concatenating all three modalities and, thus, produce a vector of size 2304 for each image chip.  We do an ablation on the size of the sample taken from the train dataset with 5, 10, 100, 250, 500, 1000, 5000, 20000 and the full dataset. The sample is random uniform across all chips on the training split within each AOI.

We therefore train (3+1) modalities $\times$ 5 AOIs $\times$ 9 train dataset sizes. We use validation data to select the overall best Random Forest configuration (50 estimators and 7 max depth), and then we measure the accuracy on the test data. This is the only time that test data is used. Observe that train data is both used to train the pretrained model and the downstream Random Forest. We repeat this procedure 10 times and report the mean value of each experiment set.

\section{Results}
\label{sec:results}

\subsection{Ablations}

\begin{figure}[h]
\centering
\caption{Accuracy on binary classification for each downstream task, AOI and modality. The x-axis is non-linear and represents the number of image chip embeddings used to train a model. Dots signal the minimum number of training image chips which with 95\% of top accuracy for each task is achieved. Observe that in the vast majority of cases, with less than 250 labeled image chip we can achieve at least 95\% of the accuracy obtained with the full training dataset of labeled images. Training dataset sizes ranges from 50K in South America to 171K in China (60\% of the total image chips in each AOI). Accuracy is measured on the full test split (20\% of data).
\label{fig:fewshots}}
\centering
\includegraphics[width=1.0\textwidth]{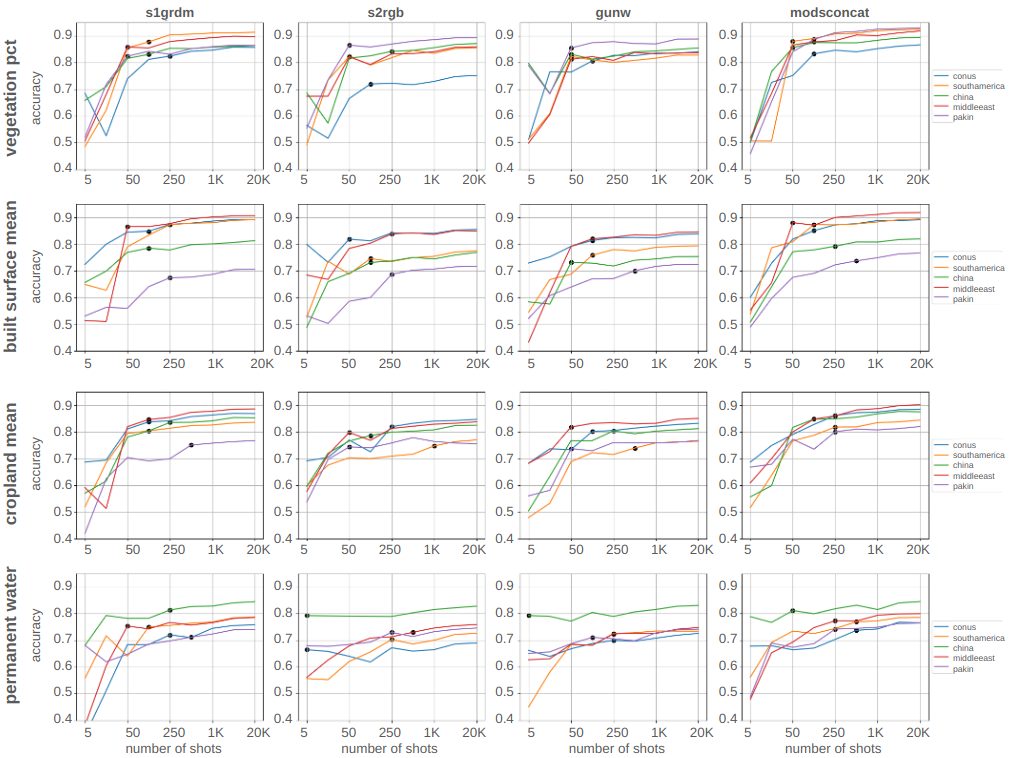}
\end{figure}

Fig. \ref{fig:fewshots} shows the overall accuracy results for our experimentation sets. Recall that we are doing a balanced binary classification task, with a different threshold in each AOI to ensure this balance, thus, reporting accuracy is straightforward. Observe that different tasks have different degrees of difficulty for different modalities. It is interesting to see that, in general, \texttt{s1grdm} embeddings perform better than the rest. Also, concatenating modalities embeddings (\texttt{modsconcat} if Fig. \ref{fig:fewshots}) seems to marginally improve overall results. We take as reference the accuracy obtained when using the full dataset, and measure how far we are from it in each experiment. Black dots in Fig. \ref{fig:fewshots} show when the experiment produces an accuracy of at least 95\% that of the one obtained with the full labeled dataset. This happens always with less than 500 image chips, and most of the times with less than 250. Considering an average training dataset size of 150K. This means that with only 0.3\% of train data (3 per thousand) we can attain 95\% of the top performance. The standard deviation was <0.05 when we used 50 or less shots, and <0.01 with larger datasets, so we did not include it in Fig. \ref{fig:fewshots} for clarity.

\label{sec:fewshots}

\begin{figure}[h]
\centering
\caption{Embeddings for each AOI and modality projected to a TSNE 2D space for visualization and colored with each downstream task label. Each dot correspond to one image chip projected to this space. These embeddings are trained and computed unsupervisedly with no label information and yet they are sensible to different land features as represented by each downstream task. Scale is logarithmic to better appreciate the value ranges of labels.
\label{fig:embeddings}}
\centering
\includegraphics[width=1.0\textwidth]{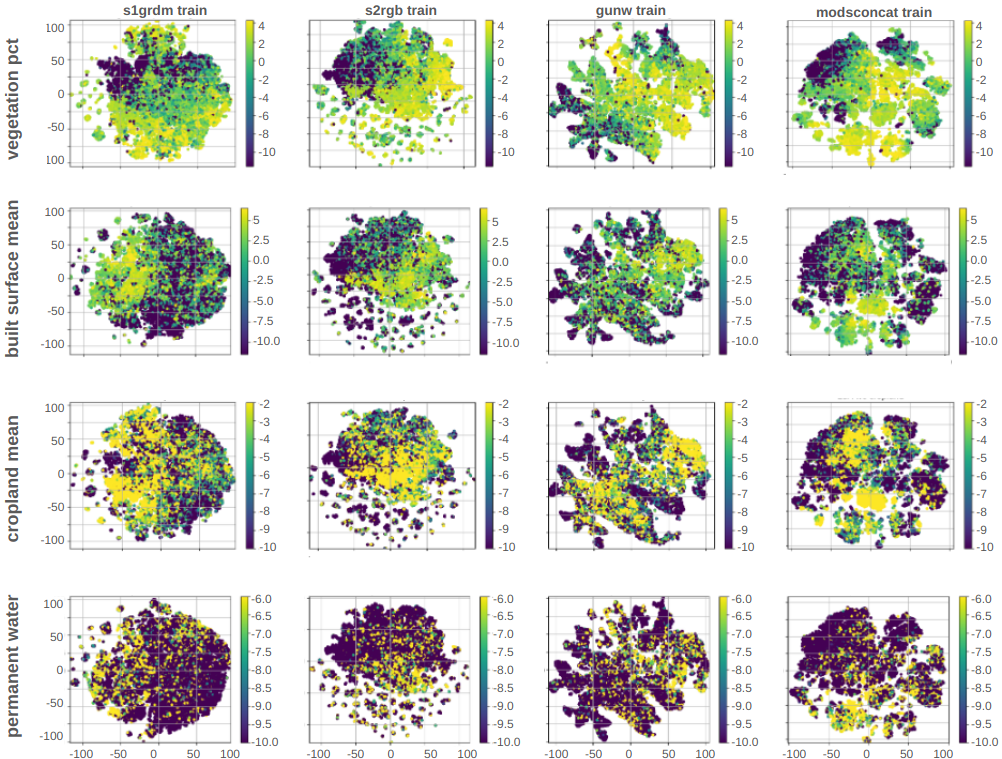}
\end{figure}

\subsection{Embeddings}
\label{sec:embeddings}

\begin{figure}[h]
\centering
\caption{Location in the \texttt{s1grdm} 2D TSNE embedding space for CONUS of two sample image chips with different vegetation percentage value. Different colouring from \ref{fig:embeddings} simply signals a different experimental run.
\label{fig:sampleimgs}}
\centering
\includegraphics[width=1.0\textwidth]{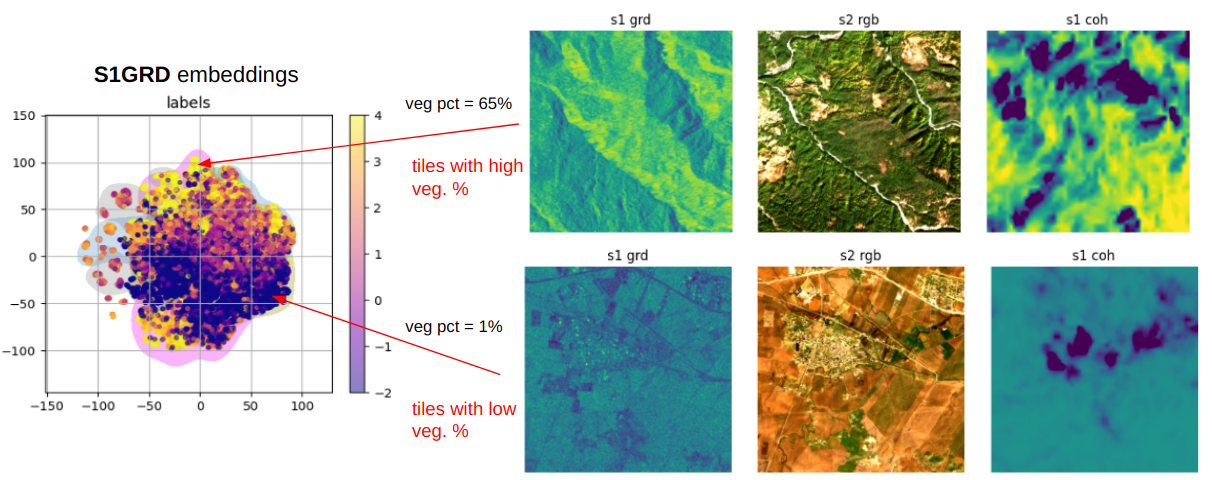}
\end{figure}

Finally, Fig. \ref{fig:embeddings} shows a 2D TSNE reduction of the embeddings obtained for each modality (columns), colored by the downstream task log mean value before thresholding for binary classification for each downstream task (rows). Observe that the labels are not used to compute neither the embeddings, nor the 2D TSNE position. And we do still get clear positional patterns where similar values of the downstream tasks cluster together. We find this significant, as it illustrates how the embeddings do capture terrain features which are useful on different downstream tasks. Although somewhat subtle, observe as well how, for the same task, different modalities separate clusters value a bit better than others. Fig. \ref{fig:sampleimgs} shows a couple of example images in the three modalities.

\section{Conclusion}
\label{sec:conclusion}
This work showed the effectiveness of multimodal models pretrained with large amounts of planetary data to reduce the number of required labeled examples in different downstream earth observation classification tasks. The reduction of the required amount of labeled data reaches the orders of 99\%.  We also run our experiments with smaller pretrained ViT architectures with 11M to 100M parameters and embeddings of size 192 and 364. Although the combined CLIP loss is usually similar to the one obtained with our 250M parameter / 768 encoding size model, the performance of the downstream tasks is degraded, even if it preserves the 95\% relative performance as described earlier. We also believe that multimodality settings such as this one allow models to leverage the complementarity or correlations of the same earth features as being observed by different sensors. This leads us to plan future work with planetary wide datasets and larger models.

\section{Acknowledgements}

%Hidden for double blind review

This work has been enabled by Frontier Development Lab Europe (\url{https://fdleurope.org}) a public / private partnership between the European Space Agency (ESA), Trillium Technologies, the University of Oxford and leaders in commercial AI supported by Google Cloud and Nvidia, developing open science for all Humankind.  L.M-F. was supported by the European Research Council (ERC) Synergy Grant “Understanding and Modelling the Earth System with Machine Learning (USMILE)” under the Horizon 2020 research and innovation programme (Grant agreement No. 855187). M. J. A. was supported by the UKRI Centre for Doctoral Training in Application of Artificial Intelligence to the study of Environmental Risks [EP/S022961/1], and additionally by Trinity Hall, Cambridge. We are also indebted to Nicolas Longépé, Carlos López-Martínez, Fabio A. González Osorio, Samuel Bancroft, Emma Hatton, Alison Lowndes, Alistair Francis, Ioanna Bouri and the rest of reviewers during 2023 FDL-Europe sprint. 

\clearpage
\bibliography{references}
\end{document}